\pdfoutput=1
\documentclass[11pt]{article}

\usepackage{ACL2023}

\usepackage{times}
\usepackage{latexsym}
\usepackage{amsmath}
\usepackage{amssymb}
\usepackage{booktabs} 

\usepackage{enumitem}
\usepackage{graphicx}
\usepackage{pifont}
\usepackage{color}
\usepackage{titletoc}
\usepackage{colortbl}
\usepackage{wrapfig}  
\usepackage{lipsum} 
\usepackage{multirow}
\usepackage{listings}

\usepackage{inconsolata}
\usepackage{ulem}
\usepackage{array}
\usepackage{amsfonts}
\usepackage[ruled,linesnumbered,noend]{algorithm2e}
\usepackage{algpseudocode}

\usepackage[T1]{fontenc}
\usepackage[utf8]{inputenc}

\usepackage{microtype}

\usepackage{inconsolata}

\newcommand{\ours}{RULE}
\newcommand{\Prob}{\mathbb{P}}

\newtheorem{Proposition}{Proposition}

\title{\ours: Reliable Multimodal RAG for Factuality\\ in Medical Vision Language Models}

\author{Peng Xia$^{1}$\thanks{$^*$Equal Contribution.}, Kangyu Zhu$^{2*}$, Haoran Li$^3$, Hongtu Zhu$^1$, \\ \textbf{Yun Li$^1$, Gang Li$^1$, Linjun Zhang$^4$, Huaxiu Yao$^1$}\\ $^1$UNC-Chapel Hill, $^2$Brown University, $^3$PolyU, $^4$ Rutgers University\\ \texttt{\{pxia,huaxiu\}@cs.unc.edu}}

\begin{document}
\maketitle
\vspace{-1em}
\begin{abstract}
The recent emergence of Medical Large Vision Language Models (Med-LVLMs) has enhanced medical diagnosis. However, current Med-LVLMs frequently encounter factual issues, often generating responses that do not align with established medical facts. Retrieval-Augmented Generation (RAG), which utilizes external knowledge, can improve the factual accuracy of these models but introduces two major challenges. First, limited retrieved contexts might not cover all necessary information, while excessive retrieval can introduce irrelevant and inaccurate references, interfering with the model's generation. Second, in cases where the model originally responds correctly, applying RAG can lead to an over-reliance on retrieved contexts, resulting in incorrect answers. To address these issues, we propose \ours, which consists of two components. First, we introduce a provably effective strategy for controlling factuality risk through the calibrated selection of the number of retrieved contexts. Second, based on samples where over-reliance on retrieved contexts led to errors, we curate a preference dataset to fine-tune the model, balancing its dependence on inherent knowledge and retrieved contexts for generation. We demonstrate the effectiveness of \ours\ on medical VQA and report generation tasks across three datasets, achieving an average improvement of 47.4\% in factual accuracy. We publicly release our benchmark and code in \url{https://github.com/richard-peng-xia/RULE}.
\end{abstract}

%%%%%%%%%%%%%%%%%%%%%%%%%%%
%Introduction
%%%%%%%%%%%%%%%%%%%%%%%%%%%
\section{Introduction}

\begin{figure}[t]
    \centering
    \includegraphics[width=\linewidth]{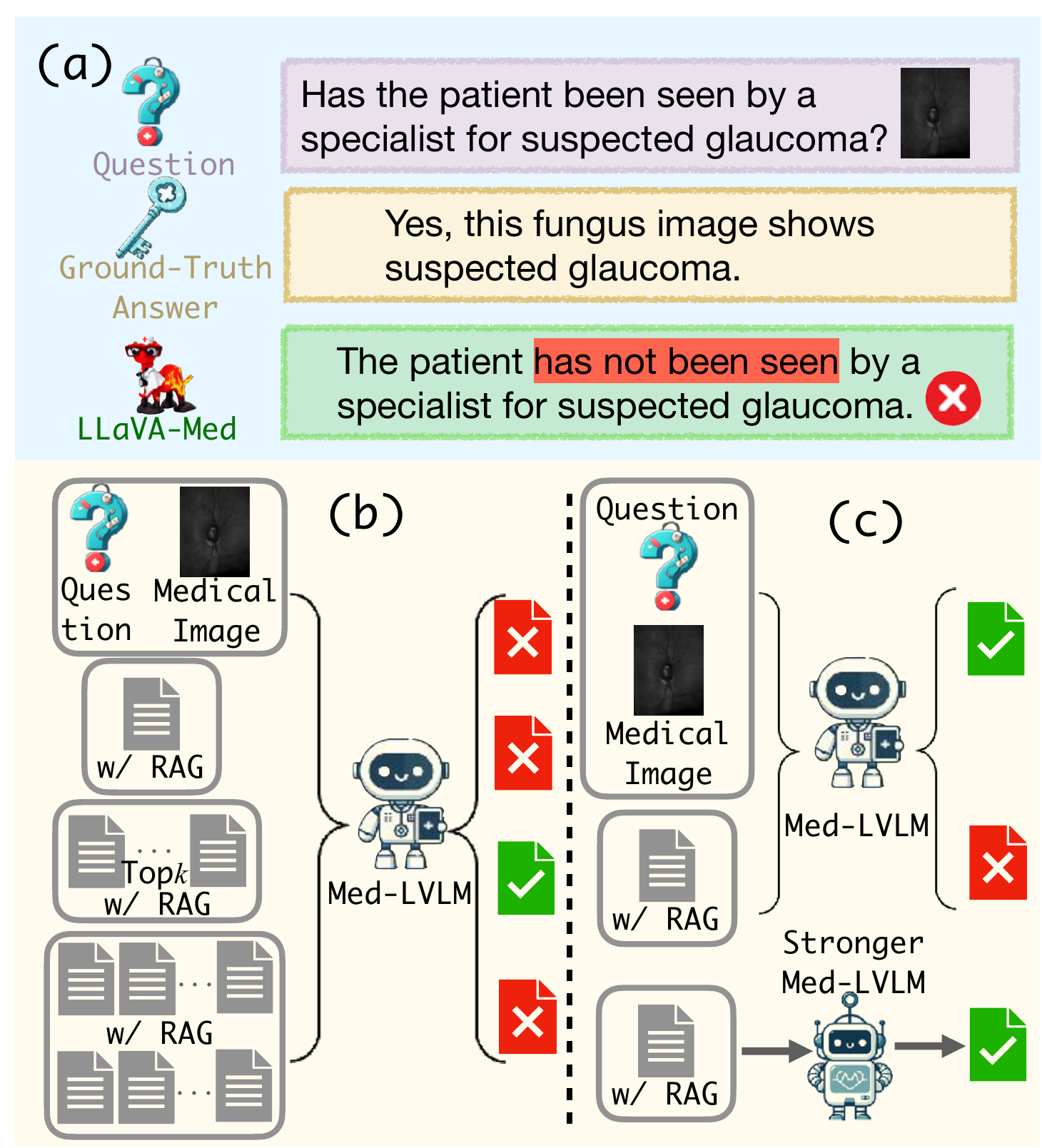}
    \caption{(a) An example of factuality issue in Med-LVLM. (b) Utilizing either too few or too many retrieved contexts as references may not provide effective guidance for the model's generation. Calibrating the number of retrieved contexts can effectively control the risk of factual inaccuracies. (c) Med-LVLMs often overly rely on retrieved contexts, leading to incorrect responses even when the original answers are correct without RAG. A stronger fine-tuned model can effectively balance its own knowledge with the retrieved contexts.}
    \label{fig:demo}
    \vspace{-2em}
\end{figure}

Artificial Intelligence (AI) has showcased its potential in medical diagnosis, including disease identification, treatment planning, and recommendations~\cite{tuauctan2021artificial,wang2019artificial,ye2021unified,xia2024generalizing,hu2024ophnet,hu2024nurvid}. In particular, the recent development of Medical Large Vision Language Models (Med-LVLMs) has introduced more accurate and customized solutions to clinical applications~\cite{li2023llava,moor2023med,zhang2023pmc,wu2023towards}. While Med-LVLMs have demonstrated promising performance, they remain prone to generating responses that deviate from factual information, potentially resulting in inaccurate medical diagnoses. This susceptibility to hallucination underscores the need for enhanced mechanisms to ensure factual alignment in critical medical applications (see an example in Figure~\ref{fig:demo}(a))~\cite{royer2024multimedeval,xia2024cares}). Such errors pose a significant risk to clinical decision-making processes and can lead to adverse outcomes.

Recently, Retrieval-Augmented Generation (RAG)~\cite{gao2023retrieval,qu2024alleviating,qu2024look} has emerged as a promising method for enhancing the factual accuracy of responses from Med-LVLMs. By integrating external, reliable data sources, RAG guides the model in producing factual medical responses, enriching its knowledge base with supplementary information. For example, RAG has been used in tasks such as visual question answering (VQA)~\cite{yuan2023ramm} and report generation~\cite{kumar2024improving,tao2024memory}. However, as illustrated in Figure~\ref{fig:demo}(b) and Figure~\ref{fig:demo}(c), directly applying RAG strategy to Med-LVLMs presents \textit{two significant challenges}: (1) A small number of retrieved contexts may not cover the reference knowledge required for the question, thus limiting the model's factual accuracy. Conversely, a large number of retrieved contexts may include low-relevance and inaccurate references, which can interfere with the model's generation; (2) Med-LVLMs may overly rely on the retrieved information. In this situation, the model might correctly answer on its own, but incorporating the retrieved contexts could lead to incorrect responses.

To tackle these challenges, we propose the \textbf{R}eliable m\textbf{U}ltimoda\textbf{L} RAG called \ours\ for M\textbf{E}d-LVLMs. First, \ours\ introduces a provable strategy for factuality risk control through calibrated selection of the number of retrieved contexts $k$, ensuring that Med-LVLMs provably achieve high accuracy without the need for additional training~\cite{angelopoulos2021learn}. Specifically, this strategy modifies the Med-LVLM through a post-processing step that performs hypothesis testing for each \( k \) to determine whether the risk can be maintained above an acceptable threshold. This process begins by calculating the \( p \)-value for each \( k \). Fixed sequence testing is then used to determine which \( k \) values can be accepted. Second, to mitigate over-reliance on retrieved knowledge, we introduce a knowledge balanced preference fine-tuning strategy. This strategy harmonizes the model's internal knowledge with retrieved contexts during medical response generation. Here, we identify samples where the model initially responds correctly but gives incorrect answers after incorporating retrieved contexts as dispreferred samples, indicating retrieval over-dependence. Conversely, ground-truth responses are considered as preferred samples. The curated preference data is then utilized for fine-tuning the preferences in Med-LVLMs.

Our primary contributions of this paper is \ours, which introduces an innovative approach to enhance retrieval-based Med-LVLMs. \ours\ not only controls factual risk by calibrating the selection of reference contexts but also balances the model's knowledge and retrieved contexts through preference fine-tuning using a curated preference dataset. Across three medical Visual Question Answering (VQA) and report generation benchmarks, including radiology and ophthalmology, our empirical results demonstrate that \ours\ effectively improves the factual accuracy of Med-LVLMs, achieving a 14.46\% improvement over the best prior methods for mitigating hallucination. In addition, empirically verify the effectiveness of the proposed components and demonstrate the compatibility of \ours. 

%%%%%%%%%%%%%%%%%%%%%%%%%%%
%Preliminaries
%%%%%%%%%%%%%%%%%%%%%%%%%%%

% - Med-LVLMs
% - preference optimization
\section{Preliminaries}
In this section, we will provide a brief overview of Med-LVLMs and preference optimization.

\noindent
\textbf{Medical Large Vision Language Models}.
Med-LVLMs connects the LLMs and medical visual modules, enabling the model to use medical images \( x_v \) and clinical queries \( x_t \) as inputs \( x \). This allows the model to autoregressively predict the probability distribution of the next token. The text output of Med-LVLMs is denoted as \( y \).

\noindent
\textbf{Preference Optimization}.
Preference optimization has achieved remarkable results in efficiently fine-tuning LLMs, significantly aligning their behavior with the goals. 
Typically, give an input $x$, a language model policy $\pi_\theta$ can produce a conditional distribution $\pi_\theta(y\mid x)$ with $y$ as the output text response. 
The recently popular DPO~\cite{rafailov2023direct} utilizes preference data achieve objective alignment in LLMs. The preference data is defined as \begin{small}$\mathcal{D}=\{x^{(i)}, y_w^{(i)}, y_l^{(i)}\}_{i=1}^N$\end{small}, where \begin{small}$y_w^{(i)}$\end{small} and \begin{small}$y_l^{(i)}$\end{small} represent preferred and dispreferred responses given an input prompt $x$. The probably of obtaining each preference pair is
\begin{small} $
    p(y_w\succ y_l)=\sigma(r(x, y_w)-r(x, y_l)),
$ \end{small}
where $\sigma(\cdot)$ is the sigmoid function. In DPO, the optimization can be formulated as classification loss over the preference data as:
\begin{equation}
\small
\begin{array}{l}
\mathcal{L}_{\textit{DPO}}(\pi_\theta; \pi_{\text{ref}}) = -\mathbb{E}_{(x,y_w,y_l) \sim \mathcal{D}} \\
\left[ \log \sigma
\left(
\alpha \log \frac{\pi_\theta(y_w | x)}{\pi_{\text{ref}}(y_w | x)}
- \alpha \log \frac{\pi_\theta(y_l | x)}{\pi_{\text{ref}}(y_l | x)}
\right) \right].
\end{array}
\label{eq:dpo}
\end{equation}
where $\pi_\theta$ represents the reference policy, which is the LLM fine-tuned through supervised learning.

%%%%%%%%%%%%%%%%%%%%%%%%%%%
%Methodology
%%%%%%%%%%%%%%%%%%%%%%%%%%%

\begin{figure*}[t]
    \centering
    \includegraphics[width=\linewidth]{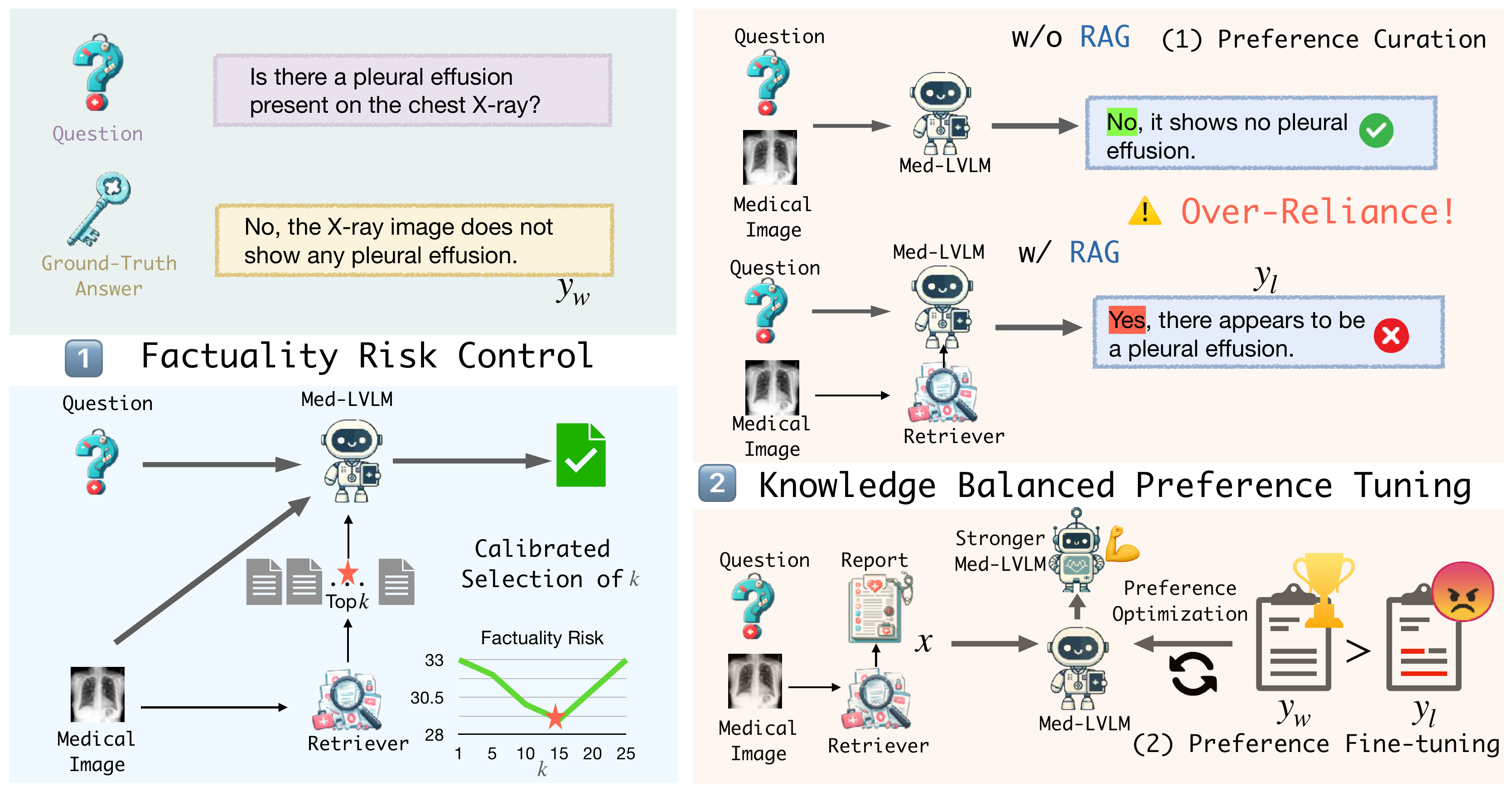}
    \caption{The framework of \ours\ comprises two main components: (1) a factuality risk control strategy through the calibrated selection of $k$; (2) knowledge-retrieval balance tuning. During the tuning phase, we initially construct a preference dataset from samples where the model errs due to excessive reliance on retrieved contexts. We subsequently fine-tune the Med-LVLM using this dataset by employing preference optimization.}
    \vspace{-1.5em}
    \label{fig:framework}
\end{figure*}

\section{Methodology}
In this section, as illustrated in Figure~\ref{fig:framework}, we will introduce \ours\ as an efficient solution for improving factuality of Med-LVLMs. Specifically, our approach consists of three main modules that work together to optimize the model's performance. First, we apply the retrieval strategy to Med-LVLMs, enhancing the model's ability to leverage retrieved information. Second, we implement a statistical method to control the factuality risk through calibrated selection of retrieved contexts. Third, we develop a preference optimization method to balance the model's reliance on its own knowledge and the retrieved contexts. Next, we will detail these three key modules in detail as follows:

\subsection{Context Retrieval for Reference}
\label{sec:retrieval}
Med-LVLMs often generate non-factual responses when dealing with complex medical images. RAG can provide the model with external knowledge as a reference, thereby effectively enhancing the factual accuracy. In the multimodal knowledge retrieval stage, \ours\ retrieves textual descriptions/reports that are most similar to the features of the target medical images. These references contain a wealth of image-based medical facts and serve to guide the generation of responses for the medical image. 

Following the design of CLIP~\cite{radford2021learning}, the retriever will first encode each image and the corresponding reports into embeddings using a vision encoder and a text encoder, respectively. Specifically, all medical images $X_{img}$ are encoded into image representations $V_{img} \in \mathbb{R}^{N \times P}$ by a vision encoder $\mathcal{E}_{img}$ (i.e., $V_{img}=\mathcal{E}_{img}(X_{img})$), where $N$ is the number of medical images that need to be retrieved, and $P$ is the dimension of the embedding. Similarly, we generate text embeddings $V_{txt} \in \mathbb{R}^{N \times P}$ for all corresponding medical reports $X_{txt}$ by applying a text encoder $\mathcal{E}_{txt}$, i.e., $V_{txt}=\mathcal{E}_{txt}(X_{txt})$. Subsequently, to adapt the general vision and text encoders to the medical domain, we fine-tune the encoders using the training data with a contrastive learning loss, defined as:
\begin{equation}
    \small
    \label{eq:clip_loss}
    \begin{aligned}
        \mathcal{L}        & = \frac{\mathcal{L}_{img}+\mathcal{L}_{text}}{2}, \\ \text{where}\;\;
        \mathcal{L}_{img}  & = -\frac{1}{N}\sum_{i=1}^{N} \log \frac{\exp(S_{i, i})}{\sum_{j=1}^{N} \exp(S_{i, j})}, \\
        \mathcal{L}_{text} & = -\frac{1}{N}\sum_{i=1}^{N} \log \frac{\exp(S_{i, i})}{\sum_{j=1}^{N} \exp(S_{j, i})},
        \vspace{-2em}
    \end{aligned}
\end{equation}
where $S \in \mathbb{R}^{N \times N}$ represents the similarity matrix between image and text modalities, calculated as: $S = \frac{V_{img}}{|V_{img}|} \cdot (\frac{V_{txt}}{|V_{txt}|})^T$, where each element $S_{i,j}$ represents the similarity between the image representation of example $i$ and the text representation of example $j$. 
Equation~\eqref{eq:clip_loss} aims to learn the representations by maximizing the similarity of text and image modalities representing the same example, while minimizing the similarity of text and image modalities representing different examples. 

After fine-tuning the image and text encoders, during inference, when faced with a target medical image $x_t$ requiring the generation of its medical report, we extract the top-$K$ similar medical reports $\mathrm{TopK}_{j \in \{1...N\}} S_{t,j}$. We then use the retrieved medical report to guide the generation of the medical report for the target medical image. with the following prompt guidance: \texttt{"You are provided with a medical image, a image-related question and a reference report. Please answer the question based on the image and report. [Question] [Reference Report] [Image]"}.

\subsection{Factuality Risk Control Through Calibrated Retrieved Context Selection}

For the RAG strategy, the top-3/5 result is typically used as a reference~\cite{gao2023retrieval}. However, it sometimes fails to encompass all relevant retrieved contexts, especially when facing the fine-grained features of medical images. Additionally, an excessive amount of retrieved contexts may introduce low-relevance and inaccurate references, which can interfere with the model's generation. 
% The quality of the retrieved contexts significantly influences the effectiveness of retrieval. 
Thus, an algorithm that can automatically determine the optimal number of retrieved contexts, based on the risk of factual errors, is particularly crucial. 

In this section, motivated by~\cite{angelopoulos2021learn}, we propose the following strategy to choose a subset $\hat\Lambda$ for the number of retrievals $k$ from a candidate set $C_K\subseteq \mathbb{N}$ such that the factuality risk $FR(k)$ can be provably controlled for any $k\in\hat\Lambda$. Specifically, first, for each $k \in C_K$, the strategy first calculates the factuality risk $FR(k)$, computed as $1-\text{ACC}(\mathcal{M}(x,(q,T_k)))$, where $x$ denotes the target medical image, $q$ denotes the question, $T_k$ means the selected top-K retrieved contexts, and $\text{ACC}(\cdot)$ measures the ratio of correct answers provided by the Med-LVLM $\mathcal{M}$ to the total number of answers. 
Next, two probabilities $p_{k1}$ and $p_{k2}$ are computed as: 
\begin{equation}
\begin{aligned}
        p_{k1} &= \exp(-nh_1(FR(k)\wedge \alpha, \alpha)),\\p_{k2} &= e\cdot \mathbb{P}(Bin(n,\alpha)\le\lceil n FR(k)\rceil),
\end{aligned}
\end{equation}
 where $h_1(a,b):=a\log(a/b)+(1-a)\log((1-a)/(1-b))$ is the Kullback-Leibler divergence between two Bernoulli distributions and $\alpha$ denotes risk upper bound. $p_{k2}$ representing the probability that, in a binomial distribution with parameters $n$ and $\alpha$, denoted by $Bin(n, \alpha)$, the observed value is less than or equal to $\lceil n FR(k) \rceil$.
Then, the minimum of these two probabilities $p_k = \min\left(p_{k1},p_{k2}\right)$ is taken.
Finally, we use any family-wise error rat (FWER)-controlling procedure, such as  Bonferroni correction \cite{van2000asymptotic} or sequential graphical testing \cite{bretz2009graphical}, to choose $\hat\Lambda$. For example, for Bonferroni correction, if $p_k$ is less than or equal to $ \delta/|C_K|$, where $\delta$ denotes tolerance level, then $k$ is added to the set $\hat{\Lambda}$. 
The proposed strategy calculates the model's factuality risk under different $k$ values, computes the corresponding probabilities using two approaches, and selects those $k$ values that meet the risk tolerance to control the overall factuality risk. 

We have the following result that ensures with probability at least $1-\delta$, the factuality risk produced is controlled by $\alpha$.
\begin{Proposition}\label{prop:risk}
Let $\alpha,\delta\in(0,1)$. 
If the training dataset $\mathcal{D}_{Med}=\{x_i, y_i, q_i\}_{i=1}^N$ is $i.i.d.$ and the output of the above algorithm $\hat\Lambda\neq\emptyset$, then $$
\Prob_{\mathcal D_{Med}}(\sup_{k\in\hat\Lambda}FR(k)\le \alpha)\ge 1-\delta.
$$
\end{Proposition}
{In practice, we calibrate the selection of $k$ on the validation sets of each dataset to minimize factuality risk. Consequently, the optimal $k$ calibrated by this algorithm can be directly used on the test sets.}

\subsection{Knowledge Balanced Preference Tuning}
\label{sec:kbpt}
In addition to selecting the optimal number $k$ of retrieved contexts, it is likely that these contents often fail to fully capture the details of every lesion or normal area in medical images. Therefore, when the retrieved contexts is inaccurate, a reliable Med-LVLM is expected to remain unaffected by the unreliable information and independently use its own knowledge to answer medical questions. However, empirically, as illustrated in Table~\ref{tab:over}, approximately half of all incorrect responses by the retrieval-augmented Med-LVLM are due to an over-reliance on retrieved contexts. This significantly affects the application of the retrieval augmented generation strategy to Med-LVLMs.

\begin{table}[h]
    \centering
    \footnotesize
    \caption{Over-Reliance Ratio (\%) of Med-LVLM with retrieval, which is the proportion of errors due to over-reliance on retrieved contexts relative to the total number of incorrect answers.}
    % \resizebox{\linewidth}{!}{
    \begin{tabular}{c|c|c}
    \toprule
        IU-Xray & FairVLMed & MIMIC-CXR \\
        \midrule
        47.42 & 47.44 & 58.69 \\
    \bottomrule
    \end{tabular}
    % }
    \vspace{-0.5em}
    \label{tab:over}
\end{table}

To address this issue, we propose a Knowledge-Balanced Preference Tuning (KBPT) strategy to mitigate over-reliance on retrieved contexts and enhance factuality in medical content generation. Specifically, we select samples \begin{small}$\mathcal{D}=\{x^{(i)}, y^{(i)}, q^{(i)}\}_{i=1}^N$\end{small} from the a separate set with samples are not used to fine-tune the retriever in Section~\ref{sec:retrieval}, where $x, y, q$ denotes input medical image, ground-truth answer and question, respectively. We identify responses $a_b=\mathcal{M}(x,q)$ where the model originally answers (i.e., $a_b=y$) correctly but gives incorrect answers $a_f=\mathcal{M}(x,(q,t))$ after incorporating retrieved contexts as dispreferred responses, as they indicate over-dependence on the retrieval. Conversely, ground-truth answers $y$ are considered preferred responses. We denote the preference dataset as \begin{small}$\mathcal{D}_o=\{x^{(i)}, y_{w,o}^{(i)}, y_{l,o}^{(i)}\}_{i=1}^N$\end{small}, where \begin{small}$y_{w,o}^{(i)}$\end{small}, \begin{small}$y_{l,o}^{(i)}$\end{small} are represented as preferred and dispreferred responses, respectively. 

Based on the curated preference data, we fine-tune the Med-LVLM using direct preference optimization. Following Eqn.~\eqref{eq:dpo}, the loss is calculated as follows:
\begin{equation}
\small
\begin{array}{l}
\mathcal{L}_{kbpt} = -\mathbb{E}_{(x,y_{w,o},y_{l,o}) \sim \mathcal{D}} \\
\left[ \log \sigma
\left(
\alpha \log \frac{\pi_\theta(y_{w,o} | x)}{\pi_{o}(y_{w,o} | x)}
- \alpha \log \frac{\pi_\theta(y_{l,o} | x)}{\pi_{o}(y_{l,o} | x)}
\right) \right].
\end{array}
\label{eq:dpoo}
\end{equation}

\begin{algorithm}
\small
    \caption{Reliable Multimodal RAG for Factuality (\textbf{\ours})}
    \LinesNumbered
    \label{ag:dpo}
    \KwIn{$\mathcal{D}=\{x^{(i)},y^{(i)},q^{(i)}\}_{i=1}^N$: Dataset; $\pi_\theta$: Parameters of the Med-LVLM; $\mathcal{D}_o$: Preference dataset; Med-LVLM: $\mathcal{M(\cdot,\cdot)}$; Retriever: $\mathcal{R(\cdot)}$; $\mathcal{D}_o$: Preference dataset.}
    \KwOut{$\pi_\text{ref}$: Parameters of the reference model.}
    $\triangleright$ \textit{Training Stage} \\
    Initialize $\mathcal{D}_o$ with an empty set\\
    \ForEach{$(x,y,q) \in \mathcal{D}$}{
        Generate retrieved contexts $t \leftarrow \mathcal{R}(x)$ \\
        Get the predictions of the model w/o retrieval $a_b \leftarrow \mathcal{M}(x,q)$ \\
        Get the predictions of the model w/ retrieval $a_f \leftarrow \mathcal{M}(x,(q,t))$ \\
        \If{$a_b=y$ and $a_f\neq y$}{
            Select the preferred response $y_{w,o} \leftarrow y $ \\
            Select the dispreferred response $y_{l,o} \leftarrow a_f $ \\
            Put $\{x,y_{w,o},y_{l,o}\}$ into $\mathcal{D}_o$\;
          }
    }
    \ForEach{$(x,y_{w,o},y_{l,o}) \in \mathcal{D}_o$}{
            Compute the losses $\mathcal{L}_o$ following Eqn.~\eqref{eq:dpoo}\\
            Update $\pi_\text{ref}$ by minimizing $\mathcal{L}_o$
        }
    $\triangleright$ \textit{Inference Stage} \\
    \ForEach{test sample $(x, q)$}{
        Select top-k retrieved contexts of calibrated algorithm $T_k \leftarrow \mathcal{R}(x)$ \\
        Get the predictions of the model w/ KBPT and retrieval $a \leftarrow \mathcal{M}(x,(q,T_k))$ \\
    }
\end{algorithm}

%%%%%%%%%%%%%%%%%%%%%%%%%%%
%Experiment
%%%%%%%%%%%%%%%%%%%%%%%%%%%

\begin{table*}[t]
    \centering
    \footnotesize
    \caption{Factuality performance (\%) of Med-LVLMs on the three VQA datasets. Notably, we report the accuracy, precision, recall, and F1 score. The best results and second best results are \textbf{bold} and \underline{underlined}, respectively. }
    \vspace{-1em}
    \resizebox{\linewidth}{!}{
    \begin{tabular}{l|cccc|cccc|cccc}
    \toprule
        \multirow{2}{*}{Models} & \multicolumn{4}{c|}{IU-Xray} & \multicolumn{4}{c|}{Harvard-FairVLMed} & \multicolumn{4}{c}{MIMIC-CXR} \\
        & Acc & Pre & Rec & F1 & Acc & Pre & Rec & F1 & Acc & Pre & Rec & F1 \\
        \midrule
        LLaVA-Med-1.5 & 75.47 & 53.17 & 80.49 & 64.04 & 63.03 & 92.13 & 61.46 & 74.11 & 75.79 & 81.01 & 79.38 & 80.49 \\ \midrule
        + Greedy & 76.88 & 54.41 & 82.53 & 65.59 & 78.32 & 91.59 & 82.38 & 86.75 & \underline{82.54} & 82.68 & 81.73 & 85.98 \\
        + Beam Search & 76.91 & 54.37 & \underline{84.13} & 66.06 & \underline{80.93} & \underline{93.01} & \underline{82.78} & \underline{88.08} & 81.56 & \underline{83.04} & \textbf{84.76} & \underline{86.36} \\
        + DoLa & \underline{78.00} & \underline{55.96} & 82.69 & \underline{66.75} & 76.87 & 92.69 & 79.40 & 85.53 & 81.35 & 80.94 & 81.07 & 85.73 \\
        + OPEAR & 70.59 & 44.44 & \textbf{100.0} & 61.54 & 71.41 & 92.72 & 72.49 & 81.37 & 69.34 & 72.04 & 79.19 & 76.66 \\
        + VCD & 68.99 & 44.77 & 69.14 & 54.35 & 65.88 & 90.93 & 67.07 & 77.20 & 70.89 & 78.06 & 73.23 & 75.57 \\
        \midrule
        \textbf{\ours\ (Ours)} & \textbf{87.84} & \textbf{75.41} & 80.79 & \textbf{78.00} & \textbf{87.12} & \textbf{93.57} & \textbf{96.69} & \textbf{92.89} & \textbf{83.92} & \textbf{87.01} & \underline{82.89} & \textbf{87.49} \\
    \bottomrule
    \end{tabular}
    }
    \vspace{-1em}
    \label{tab:com1}
\end{table*}

\begin{table*}[t]
    \centering
    \footnotesize
    \caption{Factuality performance (\%) of Med-LVLMs on the three report generation datasets. Notably, we report the average BLEU, ROUGE-L, METEOR.}
    \vspace{-1em}
    \resizebox{\linewidth}{!}{
    \begin{tabular}{l|ccc|ccc|ccc}
    \toprule
        \multirow{2}{*}{Models}  & \multicolumn{3}{c}{IU-Xray} & \multicolumn{3}{c}{MIMIC-CXR} & \multicolumn{3}{c}{Harvard-FairVLMed}  \\ \cmidrule(r){2-4} \cmidrule(r){5-7} \cmidrule(r){8-10}
        & BLEU & ROUGE-L & METEOR & BLEU & ROUGE-L & METEOR & BLEU & ROUGE-L & METEOR \\
        \midrule
        LLaVA-Med-1.5 & 9.64&12.26&8.21 &  12.11&13.05&11.16 & 18.11&11.36&10.75  \\
        \midrule
        + Greedy & 11.47&15.38&12.69 & 16.63&14.26&14.19 & 17.98&11.49&13.77 \\
        + Beam Search & \underline{12.10} & \underline{16.21} & \underline{13.17} & 16.97 & 14.74&14.43 & \underline{18.37}&\underline{12.62} & 14.50 \\
        + DoLa & 11.79&15.82&12.72 & \underline{17.11} & \underline{14.89} & \underline{14.81} & 18.26&12.51&\underline{14.51} \\
        + OPERA & 10.66&14.70&12.01 & 15.40&12.52&13.72 & 16.59&11.47&13.63  \\ 
        + VCD & 10.42&14.14&11.59 & 15.18&12.30&13.38 & 16.73&11.38&13.89  \\ \midrule
        + \textbf{\ours\ (Ours)} & \textbf{27.53} & \textbf{23.16} & \textbf{27.99} & \textbf{18.61} & \textbf{15.96} & \textbf{17.42} & \textbf{22.35} & \textbf{14.93} & \textbf{17.74} \\
    \bottomrule
    \end{tabular}
    }
    \vspace{-1em}
    \label{tab:report}
\end{table*}

\section{Experiment}
In this section, we evaluate the performance of \ours, aiming to answer the following questions: (1) Can \ours\ effectively improve the factuality of Med-LVLMs compared to other baselines and open-sourced Med-LVLMs? (2) Do all proposed components boost the performance? (3) How does \ours\ change attention weights of retrieved contexts to balance model knowledge and retrieved contexts? (4) How do different types of data or models influence DPO fine-tuning?

\subsection{Experimental Setups}
\textbf{Implementation Details}. 
We utilize LLaVA-Med-1.5 7B~\cite{li2023llava} as the backbone model. During the preference optimization process, we adapt LoRA fine-tuning~\cite{hu2021lora}. For the training of retriever, the vision encoder is a ResNet-50~\cite{he2016deep}, and the text encoder is a bio-BioClinicalBERT~\cite{alsentzer2019publicly}. We use the AdamW optimizer with a learning rate of $10^{-3}$, weight decay of $10^{-2}$ and a batch size of 32. The model is trained for 360 epochs. For more detailed information on training hyperparameters and training data, please see Appendix~\ref{sec:appendix} and \ref{sec:train}.\\
\noindent
\textbf{Baselines}. We compare \ours\ with LVLM hallucination mitigation methods that have already shown promising results in natural images, including Greedy Decoding, Beam Search~\cite{sutskever2014sequence},  DoLa~\cite{chuang2023dola}, OPERA~\cite{huang2023opera}, VCD~\cite{leng2023mitigating}. These methods manipulate the logits of the model's output tokens to enhance factual accuracy. Furthermore, we compare the performance with other open-source Med-LVLMs, including Med-Flamingo~\cite{moor2023med}, MedVInT~\cite{zhang2023pmc}, RadFM~\cite{wu2023towards}. 

\noindent
\textbf{Evaluation Datasets}.
\label{sec:dataset}
To ensure that the retrieved report content is relevant to the visual question content and to facilitate experimentation, we utilize three medical vision-language datasets, i.e., MIMIC-CXR~\cite{johnson2019mimic}, IU-Xray~\cite{demner2016preparing}, and Harvard-FairVLMed~\cite{luo2024fairclip}, encompassing radiology and ophthalmology. The training set is split into two parts: one part is used to train the retriever (Section~\ref{sec:retrieval}), and the other part is used to construct the preference dataset for KBPT (Section~\ref{sec:kbpt}). 

Additionally, we construct VQA pairs for KBPT and evaluation. Specifically, the reports from training set for preference dataset and reports from original test set are input into GPT-4~\cite{openai2023gpt4} to create closed-ended VQA data with \textit{yes} or \textit{no} answers, \textit{e.g.}, \textit{"Is there any pulmonary nodule?"}. By sampling segments from a medical report, we can generate a sequence of concise, closed-ended questions posed to the model, each with accurate answers. The questions are in \textit{yes/no} format, making it easier to analyze errors caused by over-reliance on retrieved contexts compared to open-ended questions. The detailed construction process and dataset statistics are provided in the Appendix~\ref{sec:appendix}. 

\noindent
\textbf{Evaluation Metrics}.
For Med-VQA task, we use Accuracy as the primary metric and, for detailed comparisons, we also adopt Precision, Recall, and F1 Score. For report generation task, we use BLEU Score~\cite{papineni2002bleu}, ROUGE-L~\cite{lin2004rouge} and METEOR~\cite{banerjee2005meteor} as the metrics.

\subsection{Results}
In this section, we provide comprehensive comparison results with different baseline methods and other open-sourced Med-LVLMs. 

\noindent
\textbf{Comparison with Baseline Methods}.
We present the results of a comparison between \ours\ and various hallucination reduction methods in Table~\ref{tab:com1}. According to these results, \ours\ demonstrates the best overall performance, effectively and accurately diagnosing diseases with an average accuracy improvement of 47.4\% on two tasks across all datasets. We also observe that \ours\ performs notably better on the IU-Xray and Harvard-FairVLMed compared to MIMIC-CXR. This difference is attributed to the excessive length of the reports available for retrieval in MIMIC-CXR, where overly long references tend to confuse the Med-LVLM. Even when dealing with the relatively niche ophthalmology data (i.e., Harvard-FairVLMed), \ours\ demonstrates superior results, significantly enhancing the factual accuracy of the Med-LVLM. In contrast, the performance of decoding methods is quite unstable, showing significant rates of missed or incorrect diagnoses across different datasets, as indicated by the precision and recall values.

\noindent
\textbf{Comparison with Other Med-LVLMs}.
In Table~\ref{tab:com2}, we present the comparison with different open-sourced Med-LVLMs. \ours\ demonstrates state-of-the-art (SOTA) performance across all datasets. Although the second-best model, MedVInT, outperforms other models, \ours\ achieves an average accuracy improvement of 47.4\% over it. Whether in radiology or ophthalmology, \ours\ demonstrates remarkable performance, significantly surpassing other open-source Med-LVLMs. This indicates that \ours\ is generally applicable and effective in the medical multimodal diagnosis, providing consistent improvements across various medical image modalities.

\begin{table}[t]
    \centering
    \footnotesize
    \caption{Comparison with other open-sourced Med-LVLMs. Here ``FairVLMed": Harvard-FairVLMed. }
    \vspace{-1em}
    \resizebox{\linewidth}{!}{
    \begin{tabular}{l|ccc}
    \toprule
        Models & IU-Xray & FairVLMed & MIMIC-CXR \\ 
        \midrule
        Med-Flamingo & 26.74 & 42.06 & 61.27 \\
        MedVInT & \underline{73.34} & 35.92 & 66.06\\
        RadFM & 26.67 & \underline{52.47} & \underline{69.30} \\ \midrule
        \textbf{\ours\ (Ours)} & \textbf{87.84} & \textbf{87.12} & \textbf{83.92} \\
    \bottomrule
    \end{tabular}
    }
    \vspace{-2em}
    \label{tab:com2}
\end{table}

\begin{figure*}[t]
  \centering
  \includegraphics[width=0.95\linewidth]{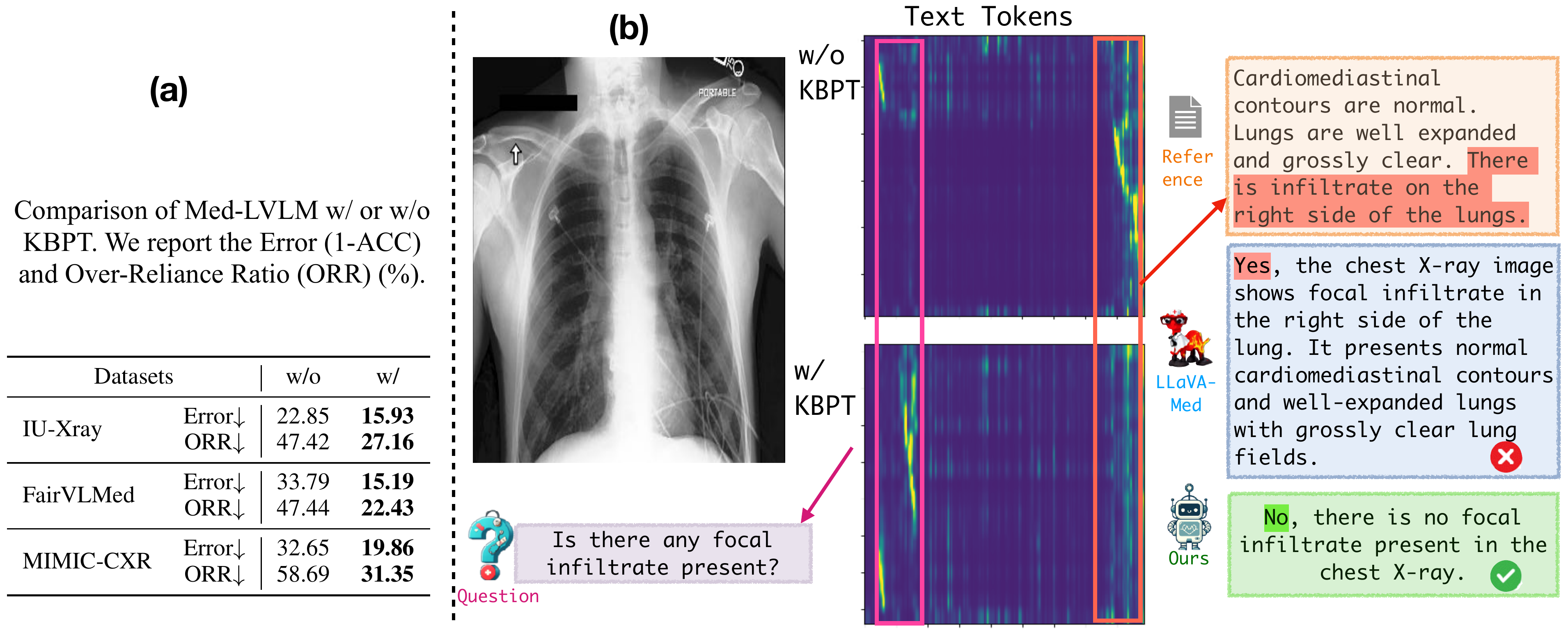}
  \vspace{-1em}
  \caption{Comparison of over-reliance metrics and attention maps. After optimizing the model with knowledge balanced preference tuning, first, (a) the Med-LVLM's error (1-acc) and over-reliance ratio significantly decrease. Second, (b) the attention scores for the latter half of the text tokens, i.e., the retrieved contexts, are significantly reduced, while the attention scores for the first half of the text tokens, i.e., the questions, have increased. It indicates that \ours\ effectively mitigates the model's over-reliance on retrieved contexts and enhances factual accuracy.}
  \label{fig:attention}
  \vspace{-1.5em}
\end{figure*}

\subsection{How Does \ours\ Improve the Performance?}
In this section, we conduct a set of analyses demonstrate how different components contribute to the performance and illustrate how \ours\ enhances overall performance, which are details as follows:

\noindent
\textbf{Ablation Studies}.
To further illustrate the effectiveness of the components of \ours, we conduct ablation experiments on three datasets. The results are shown in Table~\ref{tab:com3}. We find that the basic RAG strategy ("R") slightly improves factual accuracy on two datasets but decreases it on MIMIC-CXR. The limited retrieved contexts can not cover the fine-grained features of medical images, resulting in unstable factual accuracy improvements. With the aid of the factuality risk control strategy ("FRC"), retrieval performance see a stable increase, outperforming the original Med-LVLM. Considering the model's over-reliance on retrieved contexts, the knowledge balanced preference tuning ("KBPT") further enhances the model's reliability and significantly improves its performance. Ultimately, by combining these two strategies, \ours\ achieves optimal performance.
\begin{table}[htbp]
    \centering
    \footnotesize
    \caption{Results of ablation study. Here, ``R": retrieval; ``FRC": factuality risk control, ``KBPT": knowledge balanced preference tuning.}
    \resizebox{\linewidth}{!}{
    \begin{tabular}{l|ccc}
    \toprule
        Models & IU-Xray & FairVLMed & MIMIC-CXR \\ 
        \midrule
        LLaVA-Med-1.5 & 75.47 & 63.03 & 75.79 \\
        + R & 77.15 & 66.21 & 67.35 \\
        + FRC & 78.62 & 80.61 & 76.54 \\
        + KBPT + R & \underline{84.07} & \underline{84.81} & \underline{80.14} \\
        + \textbf{KBPT + FRC} (\textbf{Ours}) & \textbf{87.84} & \textbf{87.12} & \textbf{83.92} \\
    \bottomrule
    \end{tabular}
    }
    \vspace{-0.5em}
    \label{tab:com3}
\end{table}

\noindent
\textbf{How does \ours\ Mitigate the Issue of Over-Reliance on Retrieved Contexts?}
To better understand how \ours\ mitigates the Med-LVLM's over-reliance on retrieved contexts, we measure the Med-LVLM's error and over-reliance ratios, and visualize the text and image attention maps of the models before and after fine-tuning using a randomly selected case, as shown in Figure~\ref{fig:attention}. The quantitative results in Figure~\ref{fig:attention}(a) demonstrate the significant positive impact of \ours\ in mitigating the model's over-reliance on retrieved contexts, with the error rate and over-reliance rate decreasing by an average of 42.9\% and 47.3\%, respectively. Attention maps Figure~\ref{fig:attention}(b) illustrate the model's attention scores for text and image tokens. We find that, on the text side, the model with knowledge balanced preference tuning shows a significantly reduced focus on retrieved contexts, effectively mitigating over-reliance on such information. The model focuses more on the question and leverages its own knowledge to answer, rather than relying solely on the retrieved contexts, effectively enhancing factual accuracy. 

\noindent
\textbf{Analyzing Preference Data Type in KBPT}.
We further conduct a thorough analysis of the data types used in constructing preference data for KBPT. Three formats are considered: medical image captioning (prompted as ``Please describe this medical image"), visual question-answering (VQA), and a mixture of both. The selected data are samples where the model makes errors due to over-reliance on retrieved contexts. The results are shown in Table~\ref{tab:com4}. We observe that models fine-tuned using VQA data perform the best across all three datasets. This indicates that when retrieved contexts are incorporated into VQA questions, the Med-LVLM, through KBPT, can learn this paradigm of integrating and balancing its own knowledge with retrieved context to maximize factual accuracy. However, when the data is in the form of captioning, it may enhance the model's ability to describe medical facts, but it merely distances the model's answers from the retrieved contexts. The model fails to understand how to balance retrieval content with its own knowledge.

\begin{table}[htbp]
    \centering
    \footnotesize
    \caption{Results of models fine-tuned on different formats of data.}
    \vspace{-1em}
    \resizebox{\linewidth}{!}{
    \begin{tabular}{l|ccc}
    \toprule
        Format & IU-Xray & FairVLMed & MIMIC-CXR \\ 
        \midrule
        LLaVA-Med-1.5 & 75.47 & 63.03 & 75.79 \\
        Captioning & \underline{81.61} & 67.49 & 77.42 \\
        VQA & \textbf{84.07} & \textbf{84.81} & \textbf{80.14} \\
        Merged & 76.33 & \underline{67.96} & \underline{78.99} \\
    \bottomrule
    \end{tabular}
    }
    \vspace{-1em}
    \label{tab:com4}
\end{table}

\noindent
\subsection{Compatibility Analysis}
\begin{figure}[t]
    \centering
    \includegraphics[width=0.99\linewidth]{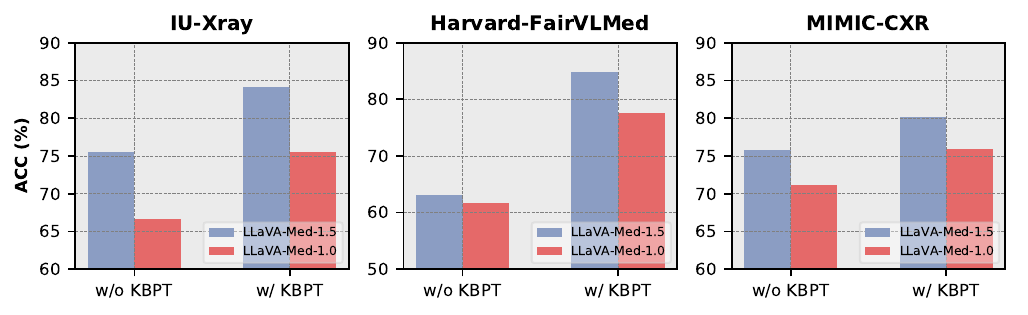}
    \caption{Results of \ours\ on different backbones. ``KBPT": knowledge balanced preference tuning.} 
    \label{fig:com4}
    \vspace{-1em}
\end{figure}

To demonstrate the compatibility of \ours, we conduct KBPT on LLaVA-Med-1.0 as well. The experimental results on three datasets are shown in Figure~\ref{fig:com4}. We find that our knowledge balanced preference tuning method demonstrates good compatibility across different models, significantly improving factual accuracy across multiple datasets. Based on LLaVA-Med-1.0, \ours\ increases accuracy by an average of 16.7\%. This indicates that \ours\ has a noticeable positive effect on mitigating over-reliance on retrieved contexts, thereby enhancing the Med-LVLM's factual accuracy.

\noindent
\subsection{Case Study}
\begin{figure}[t]
    \centering
    \includegraphics[width=0.95\linewidth]{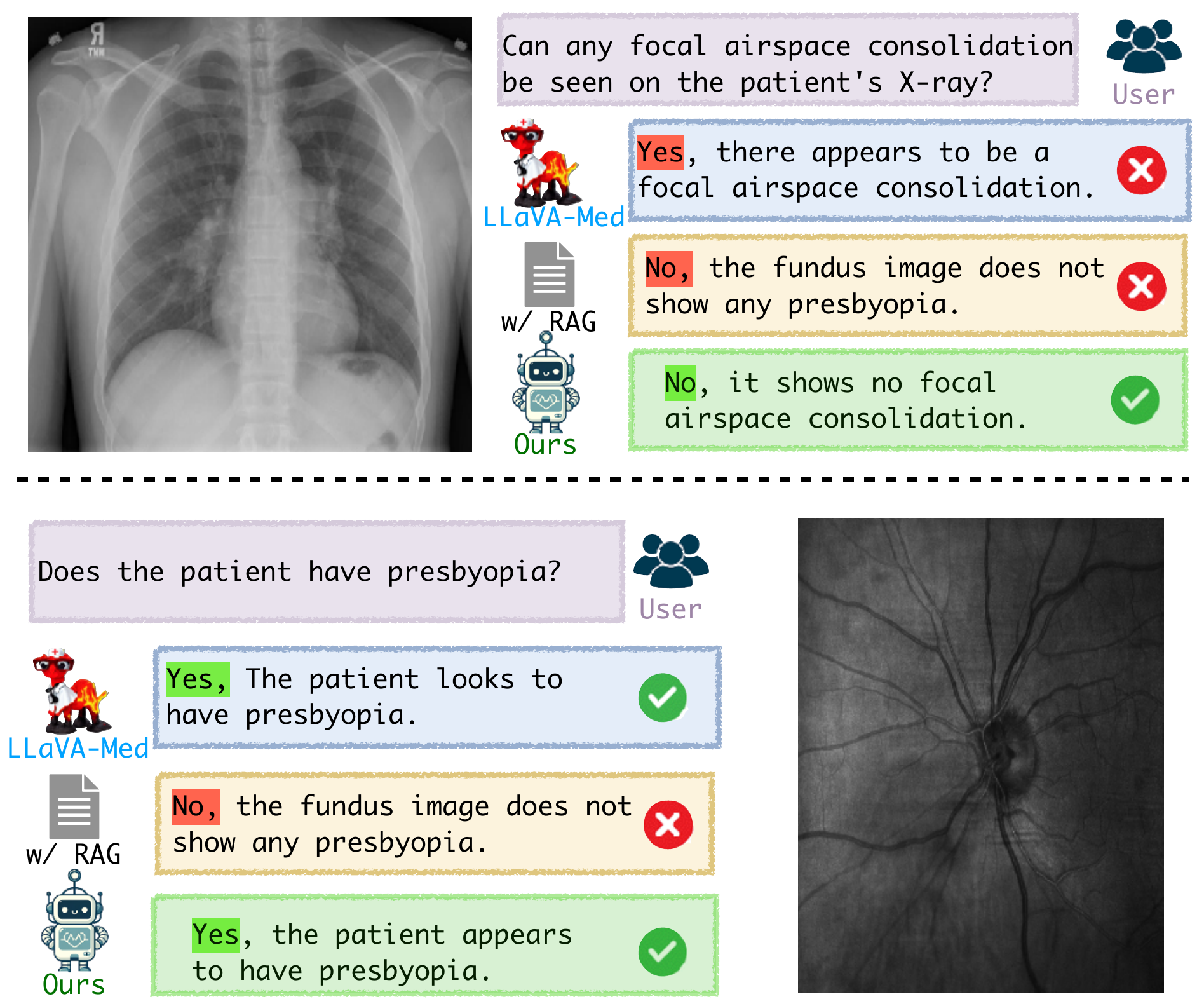}
    \caption{Illustrations of factuality enhancement by \ours\ in radiology and ophthalomology.}
    \label{fig:case}
    \vspace{-2em}
\end{figure}

Figure~\ref{fig:case} presents two representative case results, demonstrating that \ours\ can effectively enhance the factual accuracy of med-LVLMs. In case 1, LLaVA-Med provides a factually incorrect answer. After applying the RAG strategy, the model still exhibits factual issues, whereas our method effectively addresses this and improves accuracy. In case 2, LLaVA-Med initially provides a correct answer, but due to the model's over-reliance on retrieved contexts, it subsequently produces an incorrect response. \ours\ balances the weight of inherent knowledge and retrieved contexts, enhancing factual accuracy.

%%%%%%%%%%%%%%%%%%%%%%%%%%%
%Related Work
%%%%%%%%%%%%%%%%%%%%%%%%%%%
\section{Related Work}
\textbf{Factuality in Med-LVLMs}. 
The rapid development of Large Vision and Language Models (LVLMs)~\cite{liu2023visual,liu2023improved,zhu2023minigpt,alayrac2022flamingo,zhou2024aligning,zhou2024calibrated,xia-etal-2024-lmpt,xia2023hgclip} has begun to impact medical diagnosis. A series of Med-LVLMs~\cite{li2023llava,moor2023med,wu2023towards,zhang2023pmc}, represented by LLaVA-Med, have emerged, demonstrating impressive performance across various medical image modalities. However, Med-LVLMs still exhibit significant factual errors, producing medical responses that conflict with the visual medical information~\cite{xia2024cares,su2024conflictbank}. This could potentially lead to misdiagnoses or missed diagnoses. Recently, several benchmarks~\cite{royer2024multimedeval,xia2024cares} have been established to evaluate the accuracy of Med-LVLMs in tasks such as VQA or report generation. Beyond evaluating factuality, improving the factual accuracy of Med-LVLMs remains an underexplored area.

\noindent
\textbf{Retrieval Augmented Generation}. RAG has recently been recognized as a promising solution~\cite{gao2023retrieval,sun2024surf}. It enhances the model's ability to generate accurate facts by incorporating contextual information from external datasets. In medical multimodal analysis, the RAG approach has been applied to various tasks such as medical VQA~\cite{yuan2023ramm} and report generation~\cite{kumar2024improving,tao2024memory,he2024meddr}. However, in Med-LVLMs, applying RAG-based approaches overlook two critical issues: the number of retrieved contexts and whether the model overly relies on these reference. These factors can significantly affect the model's performance and may even degrade it. In \ours, we systematically address these challenges and enhance the factuality of Med-LVLMs.

%%%%%%%%%%%%%%%%%%%%%%%%%%%
%Conclusion
%%%%%%%%%%%%%%%%%%%%%%%%%%%
\section{Conclusion}
In this work, we aim to enhance the factuality of Med-LVLM by addressing two key challenges in medical RAG. 
Specifically, we first introduce a provably effective strategy for controlling factuality risk through the calibrated selection of retrieved contexts. Second, we develop a preference optimization strategy that addresses errors stemming from the model's excessive dependence on retrieved contexts, aiming to balance its intrinsic knowledge and the retrieved information. 
Experiments on three medical imaging analysis datasets demonstrate the effectiveness of \ours.
% \newpage
\section*{Limitations}
This work explores a reliable multimodal RAG method for Med-LVLMs to enhance factual accuracy. Our primary focus is on factual accuracy. Future research can explore other issues related to deploying Med-LVLMs in clinical settings, such as safety, fairness, robustness, and privacy. 

\section*{Acknowledgement}
This research was supported by Cisco Faculty Research Award.
% Currently, each dataset experiment uses a single set of data for generating retrieved contexts. In the future, it would be beneficial to build large-scale retrievable databases for more medical image modalities and to consider the efficiency of real-world deployment.

%%%%%%%%%%%%%%%%%%%%%%%%%%%
%Reference
%%%%%%%%%%%%%%%%%%%%%%%%%%%
\bibliography{custom}
\bibliographystyle{acl_natbib}

%%%%%%%%%%%%%%%%%%%%%%%%%%%
%Appendix
%%%%%%%%%%%%%%%%%%%%%%%%%%%
\newpage
\appendix

\section{Data}
\label{sec:appendix}

\subsection{Data statistics}
The quantities of all the data used are shown in Table~\ref{tab:data1} and Table~\ref{tab:data2}. It is notable to note that for training the retriever, this refers to the number of image-text pairs; for fine-tuning, it refers to the number of QA items. ``All" represents the total quantity used to construct the preference dataset, where only the samples with correct original answers that become incorrect after adding retrieved contexts are included in the training of knowledge balanced preference tuning (``KBPT").
\begin{table}[htbp]
    \centering
    \footnotesize
    \resizebox{\linewidth}{!}{
    \begin{tabular}{l|ccc}
    \toprule
    Dataset & Train (R) & All (KBPT) & Train (KBPT) \\ \midrule
    IU-Xray & 1035 & 6761 & 1579 \\
    FairVLMed & 7000 & 6271 & 2259 \\
    MIMIC-CXR & 3000 & 4951 & 1106 \\
    \bottomrule
    \end{tabular}
    }
    \caption{Data statistics of training set. Here, the number of data for the training of retriever (``R") means the number of image-caption pairs. The number of data for knowledge balanced preference tuning (``KBPT") means the number of question-answering pairs. FairVLMed: Harvard-FairVLMed.}
    \label{tab:data1}
\end{table}

\begin{table}[htbp]
    \centering
    \footnotesize
    \begin{tabular}{l|cc}
    \toprule
    Dataset & \# Images & \# QA Items\\ \midrule
    IU-Xray & 589 & 2573 \\
    Harvard-FairVLMed & 713 & 4285 \\
    MIMIC-CXR & 700 & 3470\\
    \bottomrule
    \end{tabular}
    \caption{Data statistics of test set. \# Images and \# QA items mean the number of images and QA pairs, respectively.}
    \label{tab:data2}
\end{table}

\subsection{Instructions}
\begin{table}[t]
    \centering
    \footnotesize
    \setlength{\arrayrulewidth}{0.5mm}
    \definecolor{mygray}{gray}{0.93}
    \begin{tabular}{|>{\columncolor{mygray}}p{7cm}|}
    \hline
    \textbf{Instruction [Round1]} \\
    You are a professional medical expert. I will provide you with some medical reports. Please generate some questions with answers (the answer should be yes or no) based on the provided report. The subject of the questions should be the medical image or patient, not the report. \\ Below are the given report: \\ {[REPORT]} \\
    \textbf{Instruction [Round2]} \\
    Please double-check the questions and answers, including how the questions are asked and whether the answers are correct. You should only generate the questions with answers and no other unnecessary information. \\
    Below are the given report and QA pairs in round1: \\
    {[REPORT]} \\  {[QA PAIRS R1]}\\ \hline
    \end{tabular}
    \caption{The instruction to GPT-4 for generating QA pairs.}
    \label{tab:prompt}
\end{table}
We convert the medical reports into a series of closed-ended questions with yes or no answers. To ensure the quality of the VQA data, we perform a round of self-checks using GPT-4~\cite{openai2023gpt4}. Finally, we conduct an round of manual filtering to remove questions with obvious issues or those related to multiple images or patient histories. The prompt templates used are shown in Table~\ref{tab:prompt}.

\subsection{Involved Datasets}
We utilize three open-source medical vision-language datasets, i.e., MIMIC-CXR~\cite{johnson2019mimic}, IU-Xray~\cite{demner2016preparing}, Harvard-FairVLMed~\cite{luo2024fairclip}.
\begin{itemize}[leftmargin=*]
    \item MIMIC-CXR~\cite{johnson2019mimic} is a large publicly available dataset of chest X-ray images in DICOM format with associated radiology reports. 
    % We randomly select 1,963 frontal chest X-rays along with their corresponding reports from the test set.
    \item IU-Xray~\cite{demner2016preparing} is a dataset that includes chest X-ray images and corresponding diagnostic reports. 
    % 589 frontal chest X-rays from the complete test set, along with their corresponding reports, are included in the test set of \ours. All the training set is used to train the retriever and construct the preference dataset.
    \item Harvard-FairVLMed~\cite{luo2024fairclip} focuses on fairness in multimodal fundus images, containing image and text data from various sources. It aims to evaluate bias in AI models on this multimodal data comprising different demographics. 
    % We utilize 713 pairs of retinal fundus images and textual descriptions randomly selected from the test set.
\end{itemize}

\section{Evaluated Models}
We evaluate four open-source Med-LVLMs, \textit{i.e.}, LLaVA-Med~\cite{li2023llava}, Med-Flamingo~\cite{moor2023med}, MedVInT~\cite{zhang2023pmc}, RadFM~\cite{wu2023towards}. The selected models are all at the 7B level.
\begin{itemize}[leftmargin=*]
    \item LLaVA-Med~\cite{li2023llava} is a vision-language conversational assistant, adapting the general-domain LLaVA~\cite{liu2023visual} model for the biomedical field. The model is fine-tuned using a novel curriculum learning method, which includes two stages: aligning biomedical vocabulary with figure-caption pairs and mastering open-ended conversational semantics. It demonstrates excellent multimodal conversational capabilities.
    \item Med-Flamingo~\cite{moor2023med} is a multimodal few-shot learner designed for the medical domain. It builds upon the OpenFlamingo~\cite{alayrac2022flamingo} model, continuing pre-training with medical image-text data from publications and textbooks. This model aims to facilitate few-shot generative medical visual question answering, enhancing clinical applications by generating relevant responses and rationales from minimal data inputs.
    \item RadFM~\cite{wu2023towards} serve as a versatile generalist model in radiology, distinguished by its capability to adeptly process both 2D and 3D medical scans for a wide array of clinical tasks. It integrates ViT as visual encoder and a Perceiver module, alongside the MedLLaMA~\cite{wu2024pmc} language model, to generate sophisticated medical insights for a variety of tasks. This design allows RadFM to not just recognize images but also to understand and generate human-like explanations.
    \item MedVInT~\cite{zhang2023pmc}, which stands for Medical Visual Instruction Tuning, is designed to interpret medical images by answering clinically relevant questions. This model features two variants to align visual and language understanding~\cite{wu2024pmc}: MedVInT-TE and MedVInT-TD. Both MedVInT variants connect a pre-trained vision encoder ResNet-50 adopted from PMC-CLIP~\cite{lin2023pmc}, which processes visual information from images. It is an advanced model that leverages a novel approach to align visual and language understanding.
\end{itemize}

\section{Implementation Details}
\label{sec:train}
Following the settings of CLIP~\cite{radford2021learning}, we adopt the same architecture and hyperparameters for the vision and text encoders. The vision encoder is a ResNet-50~\cite{he2016deep}, and the text encoder is a bio-bert-based model~\cite{alsentzer2019publicly}. We use the AdamW optimizer with a learning rate of $10^{-3}$, weight decay of $10^{-2}$ and a batch size of 32. The model is trained for 360 epochs. The reports available for retrieval are from the training set of the corresponding dataset. In our experiments, we apply cross-validation to tune all hyperparameters with grid search. All the experiments are implemented on PyTorch 2.1.2 using four NVIDIA RTX A6000 GPUs. It takes roughly 2.5 and 4 hours for fine-tuning CLIP and LLaVA-Med-1.5 7B, respectively.

\section{Proofs}
\textit{Proof of Proposition~\ref{prop:risk}:}
According to the definition,
{$\mathcal{M(\cdot,\cdot)}$ denotes the Med-LVLM. $\{T_k\}_{i=1}^N$ denotes the top$k$ retrieved contexts. The dataset is $\mathcal{D}_{Med}=\{x_i, y_i, q_i\}_{i=1}^N$, where $x_i$ is the target image, $y_i$ is the ground-truth answer, $q_i$ is the target question.}
By the definition of $FR(k)$,
% $$
% \text{ACC}(\cdot) = \frac{1}{N} \sum_{i=1}^{N} \mathbf{1}\{\mathcal{M}(x_i,(q_i,\{T_k\}_{i=1}^N)) = y_i\}
% $$
\begin{align*}
FR(k)=&1-\text{ACC}(\mathcal{M}(x,(q,\{T_k\}_{i=1}^N)))\\
=&1- \frac{1}{N} \sum_{i=1}^{N} \mathbf{1}\{\mathcal{M}(x_i,(q_i,\{T_k\}_{i=1}^N)) \\ =& y_i\}\\
=&\frac{1}{N} \sum_{i=1}^{N} (1-\mathbf{1}\{\mathcal{M}(x_i,(q_i,\{T_k\}_{i=1}^N)) \\ =& y_i\})
\end{align*}

Therefore, $FR(k)$ can be written as the average value of a function evaluated at each data point $(x_i, y_i, q_i)$ in $\mathcal{D}_{Med}$. 
Then, by combining Theorem 1, Proposition 1 and Proposition 2 of \cite{angelopoulos2021learn}, we finish the proof. 

\end{document}